\documentclass[letterpaper, 10 pt, conference]{ieeeconf}  

\IEEEoverridecommandlockouts                              

\usepackage{graphics} 
\usepackage{graphicx}
\usepackage{subcaption} 
\usepackage{float} 
\usepackage[font=small,skip=2pt]{caption} 
\usepackage{times} 
\usepackage{amsmath} 

\usepackage{comment}
 
\title{\LARGE \bf
Action Sequence Transfer via LLMs for Heterogeneous Environments
}

\author{Choongho Chung$^{1}$, DongHwan Shin$^{1}$, and Sung-Hee Lee$^{1,*}$
\thanks{$^{1}$ The authors are with Graduate School of Culture Technology, KAIST (Korea Advanced Institute of Science and Technology). {\tt\small \{thegenuine, sdhqaa, sunghee.lee\}@kaist.ac.kr}}
\thanks{$^{*}$Corresponding author}
}

\begin{document}



\maketitle
\thispagestyle{empty}
\pagestyle{empty}

\begin{abstract}

We present an action sequence transfer system that adaptively transfers user action sequences across different target spaces. Given an input action sequence from a source space and scene graph representations of both the source and target environments, our system predicts a corresponding action sequence in the target space by adapting to the spatial and object constraints of the new environment. To achieve this, we leverage multi-level representations of user activity to generalize actions at varying levels of abstraction. To demonstrate our system, we collect a new scene graph-based dataset derived from the Ego4D GoalStep dataset for evaluation. Results indicate that our system can generate valid action sequences even between spaces with drastically different object configurations.

\end{abstract}

\section{INTRODUCTION}

\begin{figure}[!t]
\centering
\includegraphics[width=1.0\linewidth]{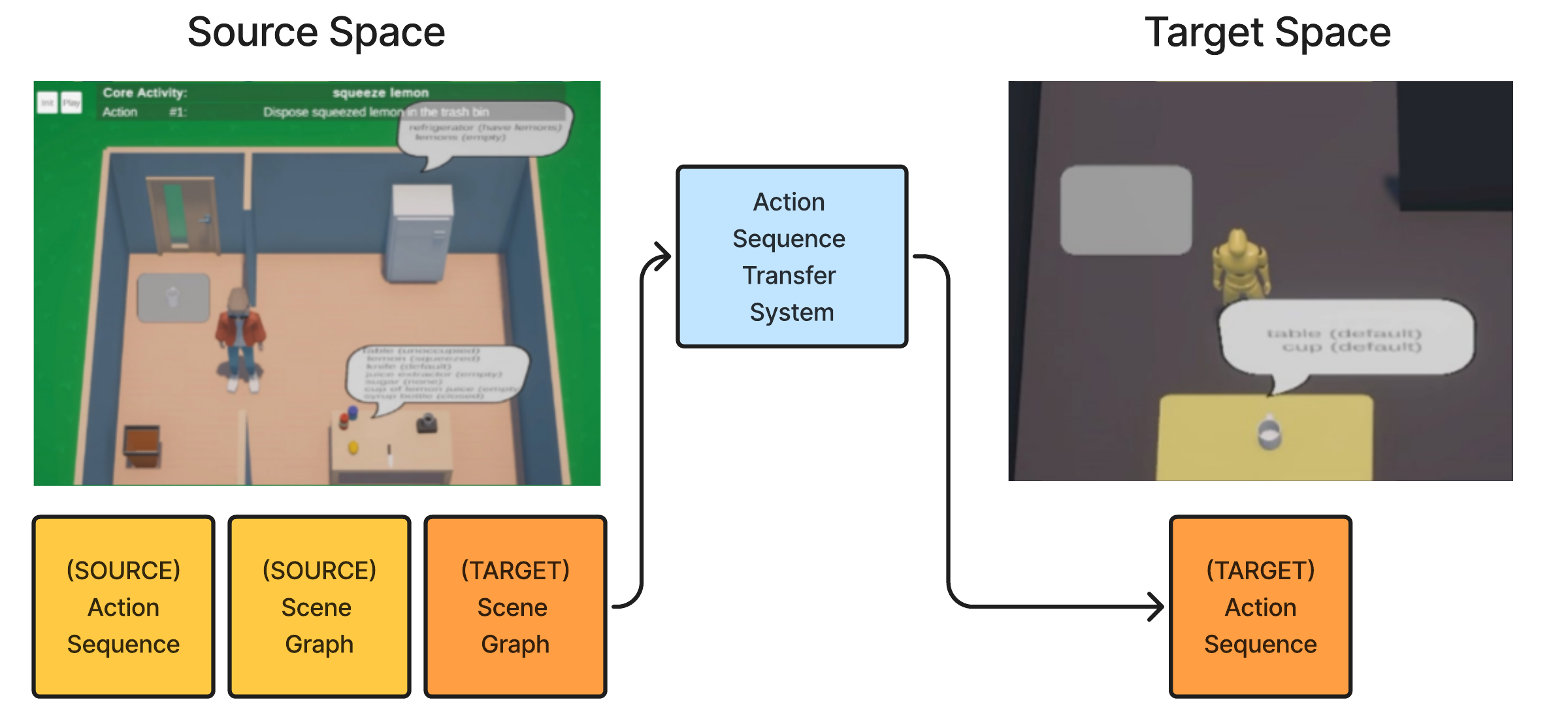}
\caption{Action sequence transfer problem between two different spaces: (a) Action sequence in source space is modified to match the environments of (b) target space, so that the overarching core activity-the goal-of the original action sequence is preserved.}
\label{fig:core_problem}
\end{figure}

In this paper, we address the problem of transferring human-demonstrated action sequences to robots operating in heterogeneous environments, where spatial layouts and object configurations differ from those in the original demonstration space (Fig. \ref{fig:core_problem}). The objective is to preserve the high-level intent of the original human actions while adapting their execution to align with the constraints and affordances of the new environment. For example, if a person boils water using a stove in one kitchen, a robot should be able to use a microwave in another kitchen where a stove is unavailable. Likewise, selection of fillings for a sandwich can switch from cheese to meat depending on available ingredients.

Rather than planning robot behavior entirely from scratch based on the goal and current environment, transferring existing action sequences allows robots to leverage the structure, preferences, and strategies embedded in human demonstrations. This approach not only reduces planning complexity and computational overhead but also enhances consistency and personalization, which are critical for effective human-robot interaction. By enabling goal-preserving adaptation of actions across diverse settings, our work contributes toward more scalable and context-aware robotic behavior.

A central challenge in this problem is maintaining the semantic fidelity of the original human activity despite differences in spatial and physical context. This requires the ability to interpret semantic relationships between actions, reason spatially about the environment, and identify the user’s underlying intent. Effective transfer also demands the prioritization of key action steps so that adapted sequences in the target environment retain the most meaningful elements of the original task. Traditional task planning or motion transfer methods \cite{wang2022predict, sayplan2023}, which often depend on proprietary or narrowly scoped datasets, struggle to generalize in such settings due to their limited semantic understanding.

\begin{figure*}[t]
\centering
\includegraphics[width=\linewidth]{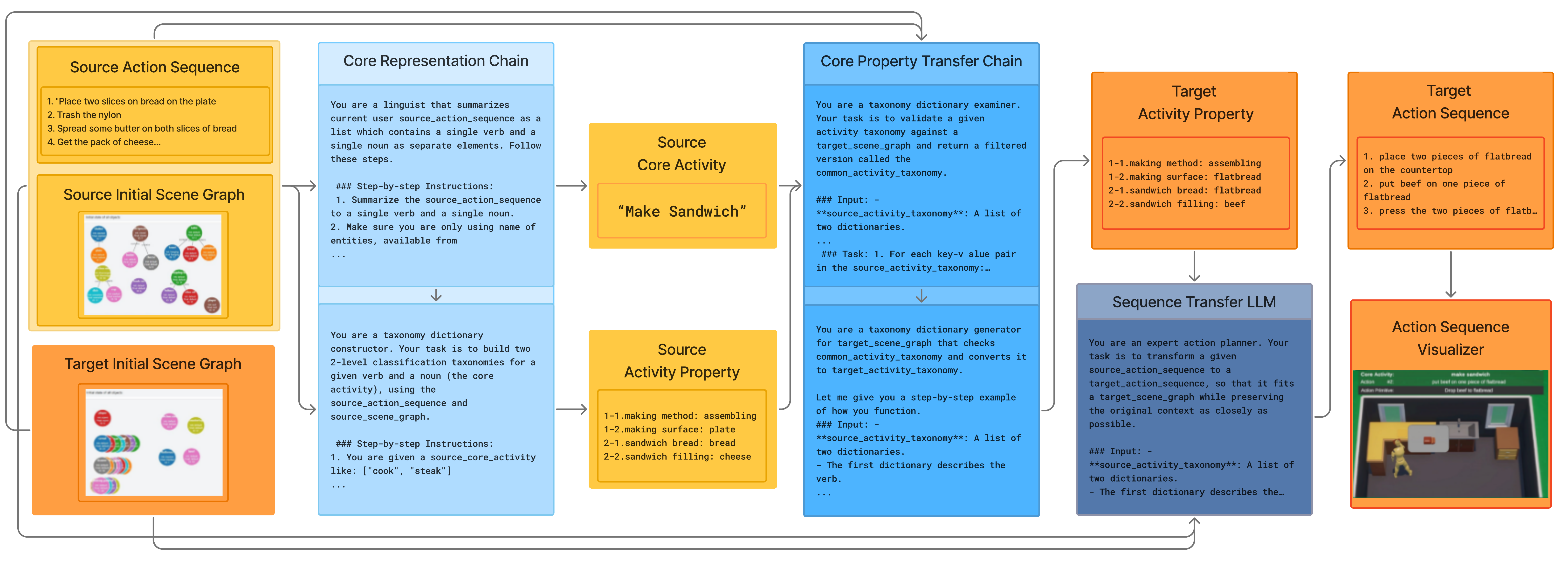}
\caption{Our action sequence transfer system predicts core activity, middle level core activity property, and the final action sequence for the target space using a sequence of LLM chains. Each LLM prompt in a chain is written to focus on generation of a single representation format. We visualize the resulting target action sequence after dividing each target action into executable action primitives for animation. }
\label{fig:system_overview}
\end{figure*}

To address this, we propose a large language model (LLM)-based action sequence transfer system (Fig. \ref{fig:system_overview}) that infers high-level user intent and rewrites action sequences to suit new environments. 


Our method leverages the broad world knowledge and abstraction capabilities of LLMs, by prompting our system to describe detailed user actions into higher level descriptions on user objectives. These middle or high level descriptions are then employed in output sequence generation steps, to act as guidelines for generating the output sequence. Our approach allows for flexible restructuring of the input action sequence by enabling exploration of substitute objects even when the target space lacks direct object correspondences.

As an additional significant contribution, we construct the Goalstep-Spatial dataset, derived from the Ego4D Goalstep dataset \cite{grauman2022ego4d, song2023ego4d}. This new dataset is used both for prompt augmentation and for the evaluation of our system, and encompasses over 44 hours of video data, annotated with scene graph representation and corresponding activity description. It enriches the goal prediction phase of our pipeline by providing contextually relevant information through Retrieval-Augmented Generation (RAG) \cite{lewis2020retrieval}, enabling more accurate and semantically grounded action sequence transfer. The dataset and source code will be released for research purposes.

\section{Framework Formulation}

\section{RELATED WORKS}

\subsection{LLM Agents for Goal-Conditioned Action Planning}

Recent advances in deep learning and LLMs have enabled autonomous agents to generate temporally structured action plans from high-level user goals in virtual environments \cite{puig2018virtualhome, li2022interactive, Su23LangGuidedProg, wang2023describe, singh2024anticipate}. Real-world applications involving robotic manipulators \cite{singh2023progprompt, gupta2024action} often rely on multi-modal input data, such as visual embeddings \cite{wu2023embodied}. These models leverage hierarchical representations of user-environment interactions and benefit from rich datasets that annotate multi-step, goal-oriented tasks \cite{luo2021moma, luo2022moma, grauman2022ego4d, durante2024few}.

LLM-based planners can operate in few-shot or zero-shot settings, and some systems augment inference with planning validation or environmental feedback, either in virtual environments \cite{wang2023describe, wang2024jarvis} or in real-world settings \cite{rana2023sayplan, singh2023progprompt, gu2024conceptgraphs}. However, most of these systems are grounded in static environments and lack mechanisms for adapting plans to different spatial contexts. To address this limitation, our system adapts to varying spaces by planning for user goals and identifying substitutes for missing source-space objects in the target space.

\subsection{LLM Comprehension of Hierarchical Concepts}

Many LLM-based planners and autonomous agents rely heavily on the conceptual knowledge embedded within the models themselves. Understanding how LLMs represent lexical and conceptual knowledge has been the focus of extensive research. Although the full nature of LLM behavior is not yet fully understood, studies have shown that LLMs partially encode conceptual knowledge in vectorized form, allowing semantic domains to be probed using linear methods \cite{park2311linear}.

Emerging evidence also indicates that LLMs capture hierarchical relationships between concepts via geometric structures within their embedding space \cite{park2024geometry}. This capability supports their effectiveness in tasks such as taxonomy enrichment, taxonomy construction, and lexical entailment \cite{moskvoretskii2024large, moskvoretskii2024taxollama}. These findings suggest that hierarchical concept representations offer a more stable and interpretable foundation for leveraging LLM domain knowledge in classification and semantic reasoning tasks. Concise user action descriptions, especially those incorporating multi-level user goals or intentions, can effectively condition the LLM to predict lower-level action plans aligned with the overarching goals.

\subsection{Employing Goal Prediction in Sequence Anticipation}

Long-Term Anticipation focuses on inferring user goals from behavioral trajectories over extended durations \cite{mittal2024can}. Zhao et al. \cite{zhao2023antgpt} combine bottom-up observation with top-down reasoning to address the inherent uncertainty in long-horizon predictions. These approaches use past time-series data of user actions to predict higher-level user intentions that guide the execution of those actions. The inferred high-level intentions can then serve as priors for generating the necessary future actions required to complete a given task.

In contrast, our work starts with a complete action sequence and restructures it to suit a new target space. We do this by first predicting the user's high-level intention, which then conditions the LLM to reconstruct a semantically aligned action sequence that conforms to the spatial and object constraints of the target environment.

\subsection{Spatial Retargeting and Scene Adaptation}
Prior works on remote collaboration emphasizes motion retargeting and spatial placement between environments \cite{yoon2020placement, li2021synthesizing}. These methods attempt to translate user behavior to virtual avatars via geometric mappings or subspace transformations \cite{keshavarzi2022mutual, choi2023online, wang2022predict}. However, they often lack semantic understanding of user tasks, making them less effective for generalizing across scenes with different object inventories and spatial configurations.

Building on prior findings regarding LLM comprehension of multi-level concepts and goal-conditioned action prediction, we extend this knowledge to our unique challenge of transferring user action sequence to heterogeneous environments. Our approach infers user intentions at multiple levels of abstraction and employs them in generating an action sequence tailored to the target space. This enables a robot avatar to replicate the user's activity while preserving semantic intent across different environmental contexts.

\section{Data Collection}

\begin{figure}[t]
\centering
\includegraphics[width=1.0\linewidth]{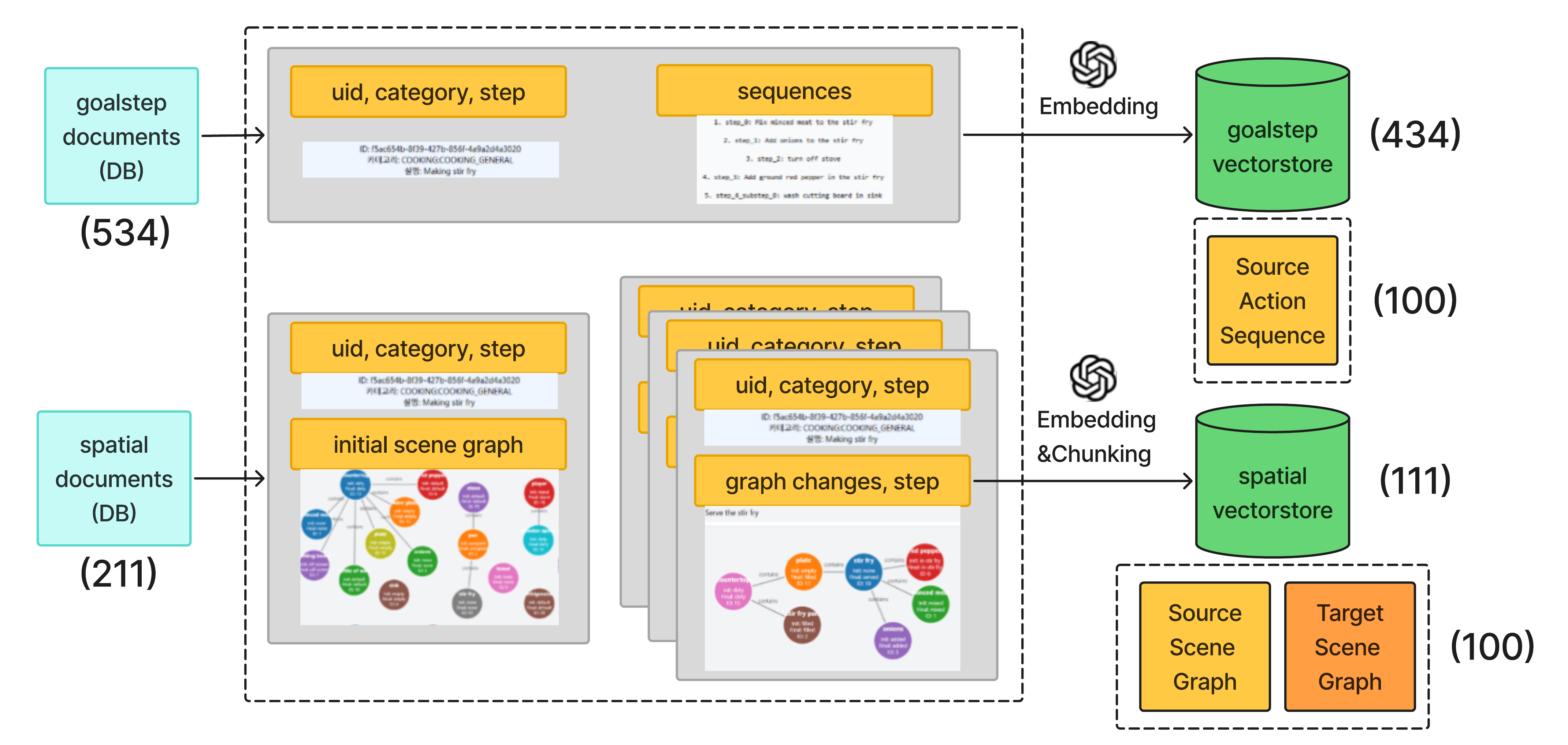}
\caption{Our dataset consists of two components: (1) a reformatted Ego4D Goal-Step dataset, and (2) a dataset containing initial scene graphs along with corresponding scene graph changes aligned with its sub-steps as actions. We select 100 pairs of Goal-Step sequences and their corresponding initial spatial scene graphs as the evaluation dataset. The remaining data are stored in a vector store and used for prompt augmentation during the core activity prediction step.}
\label{fig:data_overview}
\end{figure}

\subsection{Data Collection}

\begin{table}[t]
\centering
\resizebox{\columnwidth}{!}{%
\begin{tabular}{|l|c|c|c|c|c|c|}
\hline
\multicolumn{7}{|c|}{\textit{scenario count for spatial dataset annotation}} \\
\hline
\textbf{Scenario} & \textbf{Cooking} & \textbf{Baking} & \textbf{Laundry} & \textbf{Talking} & \textbf{Eating} & \textbf{Total Time} \\
\hline
All Dataset  & 166 & 32 & 28 & 9 & 5 & 44.0 hours \\
Evaluation   & 83  & 6  & 9  & 8 & 2 & 22.6 hours \\
\hline
\end{tabular}
}
\caption{Scenario count (overlap is allowed) statistics for the annotated spatial dataset and the evaluation dataset splits.}
\label{table:dataset_stats}
\end{table}

We collect a paired dataset consisting of action sequences and scene graph annotations for multiple spaces. Our dataset consists of two primary components: user goal-step data and spatial data (Fig. \ref{fig:data_overview}). 

\vspace{0.5em}
\textbf{Goal-Step Dataset:}
The goal-step data is derived from the Ego4D Goal-Step dataset \cite{grauman2022ego4d, song2023ego4d}, which provides action step annotations for various egocentric videos and scenarios, along with corresponding user goals. Each video in the original Ego4D dataset captures a single user activity. Our Ego4D Goal-Step further segments this activity into action steps and finer-grained sub-steps, yielding a multi-level description of the task. We reformat the test split of this dataset—comprising 534 annotated videos spanning over 300 hours of user activity, so that it includes only the video id, goal descriptions, goal categories, steps, and sub-steps acting as the action sequence for the single video data.

\vspace{0.5em}
\textbf{Spatial Dataset:}
Our spatial annotation data consists of two types of scene graphs: an initial scene graph representing the starting state of the environment, and step-specific graphs representing changes occurring at each sub-step of the corresponding Goal-Step data. Each node in the graph corresponds to an object in the environment and encodes the object’s name, ID, initial state, and final state. For each sub-step, the corresponding scene graph captures the participating objects and their state changes.
 
We collect scene graph data in two tiers of ``Gold Data” and ``Silver Data,” referring to a previous work on annotation strategy \cite{lim2024vlr}. Gold Data is created by manually labeling videos by cross checking the goal-step annotations from the Ego4D dataset. In total, we generate 65 instances of Gold Data. Silver Data, consisting of 146 annotations, is generated by following the annotation strategy in LLaVA \cite{liu2023llava}. We extract snapshots at fixed time intervals, composite them into a single image, and feed it to LLaVA 1.6 model to produce initial scene graph and its corresponding changes. Both types of graphs are manually refined to ensure temporal consistency with the associated goal-step annotations. In total, six annotators from a local university (3 male, 3 female; age: M = 28.3, SD = 3.0) created 211 scene graph annotations spanning 44 hours of video data. (Tab. \ref{table:dataset_stats}).

\vspace{0.5em}
\textbf{Evaluation Dataset:}
From the collected data, we select 100 initial scene graphs from the spatial annotations along with their corresponding Goal-Step action sequences to serve as our evaluation dataset. The remaining Goal-Step and spatial annotations are embedded using text-embedding-ada-002 from the OpenAI API and stored in separate FAISS databases \cite{douze2024faiss}. These databases are queried during retrieval-augmented generation to provide relevant examples for user intention prediction.

\subsection{Formulation of Space Pairs in Test Data}

\begin{figure}[t]
\centering
\includegraphics[width=1.0\linewidth]{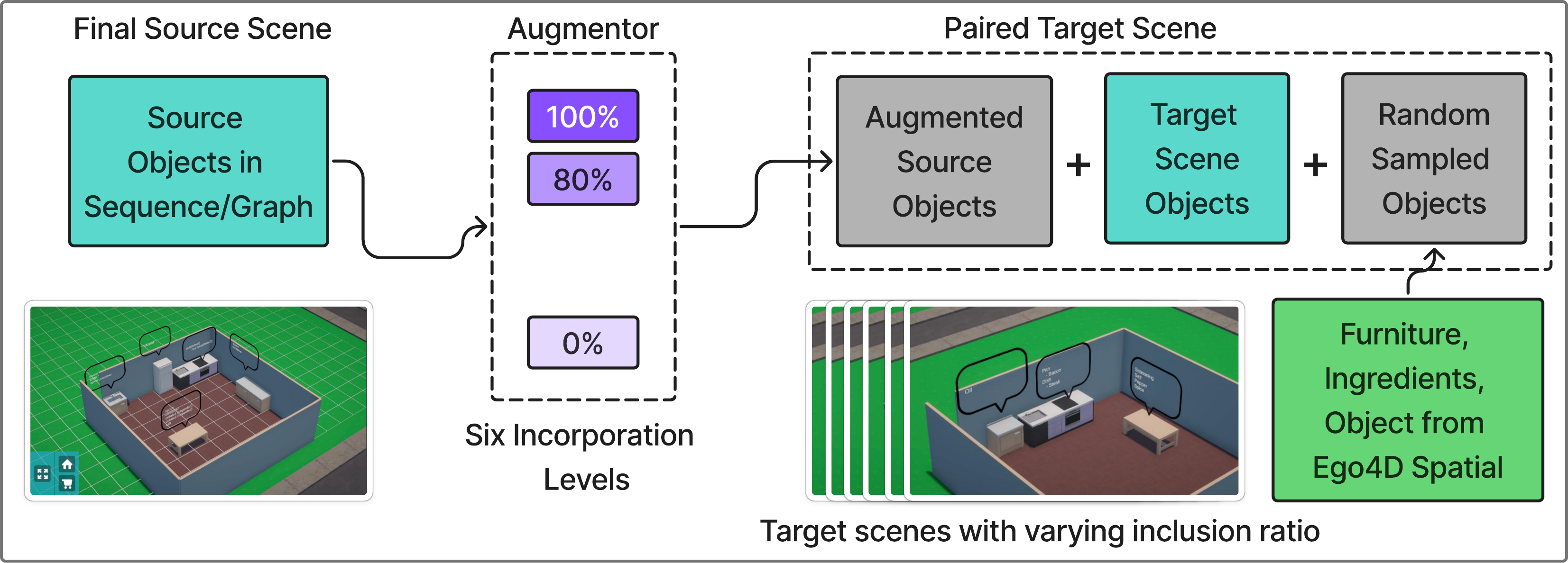}
\caption{A two-step target scene augmentation procedure is used to generate unique source–target space test pairs for system input.}
\label{fig:augmentation_overview}
\end{figure}

To evaluate our system, we pair each of the 100 selected Goal-Step sequences with corresponding source and target initial scene graphs. The target scene graph in these pairs are further augmented to generate multiple variations with different levels of similarity. This process allows us to assess the framework’s sensitivity to spatial differences while increasing the evaluation dataset size to 600 pairs (Fig. \ref{fig:augmentation_overview}).


The augmentation process consists of two steps. First, we vary the proportion of source objects (objects in the source action sequence and source scene graph) incorporated into the target scene graph across six levels: 0\%, 20\%, 40\%, 60\%, 80\%, and 100\%. Second, we introduce additional objects sampled from a lexicon of object names compiled from the Ego4D Spatial dataset. This lexicon is further categorized into ingredients, tools, and furniture. For each of the 100 selected pairs, we randomly sample 30 ingredients, 10 tools, and 2 pieces of furniture. These sample sizes are heuristically chosen to avoid excessive clutter in the target space. The sampling process serves two purposes: 1) ensuring a more uniform distribution of common objects across source and target spaces, and 2) enriching the scene graphs to enable flexible substitutions with alternative objects during task execution.

The object overlap after final augmentation is represented by the inclusion ratio, defined as the proportion of objects in the source action sequence that are also present in the target initial scene graph.

\subsection{Problem Formulation}

Our system takes as input an action sequence $S=[s_0,...,s_n]$, consisting of individual actions $s_i$, and an initial scene graph $G$ extracted from the source space, along with a corresponding initial scene graph $G_t$ representing the target space (Fig. \ref{fig:core_problem}). The initial scene graph nodes represent distinct objects in the environment while its edges capture possessive relationships between them. The objective is to predict target sequence $S_t=[s'_0,...,s'_n]$, which executes the intended goal of the original action sequence while matching the configuration of the target space.

\subsection{Multi-Level Representations of Human Activity}
\label{sec:multi-level_rep}
To generalize information from the action sequence and scene graph into a more transferable format, we leverage a core activity representation $C = [n, v]$, defined as a verb–noun pair that encapsulates the user’s primary objective.

The core activity promotes generalization, but it may struggle to represent the finer semantic distinctions in complex tasks. To address this problem, we introduce a middle level representation of core activity property $P$. This consists of two dictionaries, each corresponding to the verb and noun components of the core activity, respectively. Each dictionary holds two key–value pairs, where the keys denote relevant semantic properties and the values specify the associated objects or concepts in the environment. For example, in Fig. \ref{fig:system_overview}, the core activity property is defined by assigning two key–value pairs each for the verb “make” and the noun “sandwich”.

Informed by recent research on hierarchical concept encoding in LLMs \cite{park2311linear, park2024geometry, moskvoretskii2024taxollama}, we organize these properties by importance, with higher-level or more influential attributes listed first. We intentionally limit our core property to use two key-value pairs to balance semantic richness with the risk of over-constraining the system if deeper hierarchies are imposed.

Together, these multi-level representations—core activity, core properties, and the action sequence—form a hierarchy that governs action sequence adaptation.  Initial scene graphs constrain feasible goals for each environment, while the inferred core activity filters which properties and actions are necessary for adaptation.

\subsection{Action Sequence Transfer Pipeline}
Our action sequence transfer framework consists of three sequential LLM modules based on GPT-4o \cite{achiam2023gpt}: (1) the core representation predictor chain, (2) the core property transfer chain, and (3) the sequence transfer LLM (Fig. \ref{fig:system_overview}). We leverage three levels of activity representation formats in our framework and, to minimize formatting errors, assign a dedicated LLM model to each type of output format. We plan to release complete prompt templates and additional experimental details in our public code repository for reproducibility.

We test our action sequence system by setting up two heterogeneous spaces (Fig. \ref{fig:visualization_example}), whose spatial initial scene graph on objects is sampled from the test data. Each of the household is set to have different spatial layout on geometry, and room configurations. The input action sequences $S$ mainly consist of scenarios on cooking or manipulation of cooking appliances. 

\vspace{0.5em}
\textbf{Core Representation Predictor Chain:} The core representation predictor chain predicts the core activity and its properties in succession as in Eq. \eqref{eq:c}; In the first step, an LLM model $\Phi_c$ summarizes the source space action sequence $S$ and scene graph $G$ into a core activity $C$. We also employ RAG for prompt augmentation to retrieve similar examples from unseen data to provide additional guidance for core activity inference. To concisely conceptualize the input action sequence, we prompt the model to generate a single verb noun pair, by only using words available from the input initial scene graph $G$. 

Prompts for the LLM models are organized into four components: task, input, step, and output format. Each prompt begins by defining the model’s role or task, followed by concise description and format of each inputs. The step component specifies procedural tasks for processing inputs and generating predictions, while the format component enforces consistent and well structured results. Depending on task complexity, certain components may be merged or omitted; for example, in the case of models in the core representation chain, the input sections are excluded to emphasize the task specification. 

\begin{align}
C &=\Phi_c(p_\text{rag}, S, G), \label{eq:c} \\ %
p_\text{rag} &= f_{gs}(S) \oplus f_s(G)
\label{eq:rag}
\end{align}

The augmented prompt $p_\text{rag}$ combines retrieved values from two FAISS database retrievers constructed in data collection: $f_{gs}$ for the goal-step database and $f_s$ for the spatial database. For the goal-step retriever, the source action sequence acts as a query $S$, and cosine similarity is used to return documents containing the goal description, goal category, steps, and sub-steps of the top 3 most similar examples. Similarly, the spatial data retriever is queried with the source initial scene graph $G$ to retrieve the top 3 most similar initial scene graphs along with their corresponding graph changes. The information in the retrieved data provides a list of relevant relationships between specific actions, goals, and necessary objects.

The second LLM model $\Phi_p$ in the chain infers the key property of the referred core activity as described in Sec. \ref{sec:multi-level_rep}. We prompt the LLM to generate keys that either describe the noun or the verb of the core activity to obtain a close semantic link between the two representations.
\begin{equation}
    P=\Phi_p(p_\text{rag}, S, C, G)
\end{equation}

For a sequential activity of making ``steak", a core activity of ``cook steak" can be predicted. The core property $P$ is predicted by the module to add more detailed information on the core activity. For example, key-value pairs of ``cooking method: roasting", and ``cooking vessel: pan" might be inferred for ``cooking". Similar reasoning applies for ``steak" with examples such as ``steak seasoning: pepper", ``steak garnish: onion" when such objects are detected in the input actions.

\textbf{Core Property Transfer Chain:} 
After predicting the core representations, the core property transfer chain identifies suitable property values that fit the target environment, while preserving the core activity and property key names. The target core property prediction is divided into two steps of absent value identification step and value replacement step, which are separated to maximize the possibility of correctly identifying absent source space objects in the target space.

The first LLM $\Phi_{cp}$, prompted by $p_{cp}$, evaluates whether the core property values from the source scene can be maintained in the target environment. The prompt is designed to either preserve or clear the values of the core property dictionaries when no identical procedures or items are found in the target initial scene graph $G_t$, resulting in a common property $P_{cp}$. For a target space without ``pepper" and ``pan", this leads to the corresponding key values to be filled with ``empty" values.
\begin{equation}
P_{cp} = \Phi_{cp}(p_{cp}, S, C, P, G, G_t)
\label{eq:cp}
\end{equation}

The second LLM $\Phi_{tp}$, prompted by $p_{tp}$, fills in any empty values of the core property dictionaries with the closest alternative concepts or objects available in the target space, producing the target property $P_t$. When substitutes for empty objects such as ``pot" for ``pan" and ``salt" exists in target spaces, such values are used to replace the empty values for the core property.
\begin{equation}
P_{t} = \Phi_{tp}(p_{tp}, S, C, P, P_{cp}, G, G_t)
\label{eq:tp}
\end{equation}

\textbf{Sequence Transfer LLM:} The final sequence transfer LLM $\Phi_{ts}$ takes all the available input and intermediate representations for both the source and the target spaces to generate the final action sequence.
\begin{equation}
S_{t}=\Phi_{ts}(p_{ts}, S, C, P, P_t, G, G_t)
\label{eq:st}
\end{equation}

This final LLM, prompted by $p_{ts}$, iterates over each step in the original action sequence and apply necessary modifications based on the core activity and the target space core properties, while strictly using only objects available in the target space. The resulting modified action sequence is then validated for executability and further refined to produce the final target action sequence. For the ``cooking steak" scenario, the resulting sequence retains the core activity while, the core property and the action sequence involves use of pot and salt as substitutes for the unavailable objects from the source scene.


\section{EVALUATION}

\subsection{Result Examples}

\begin{figure}[h]
\centering
\includegraphics[width=1.0\linewidth]{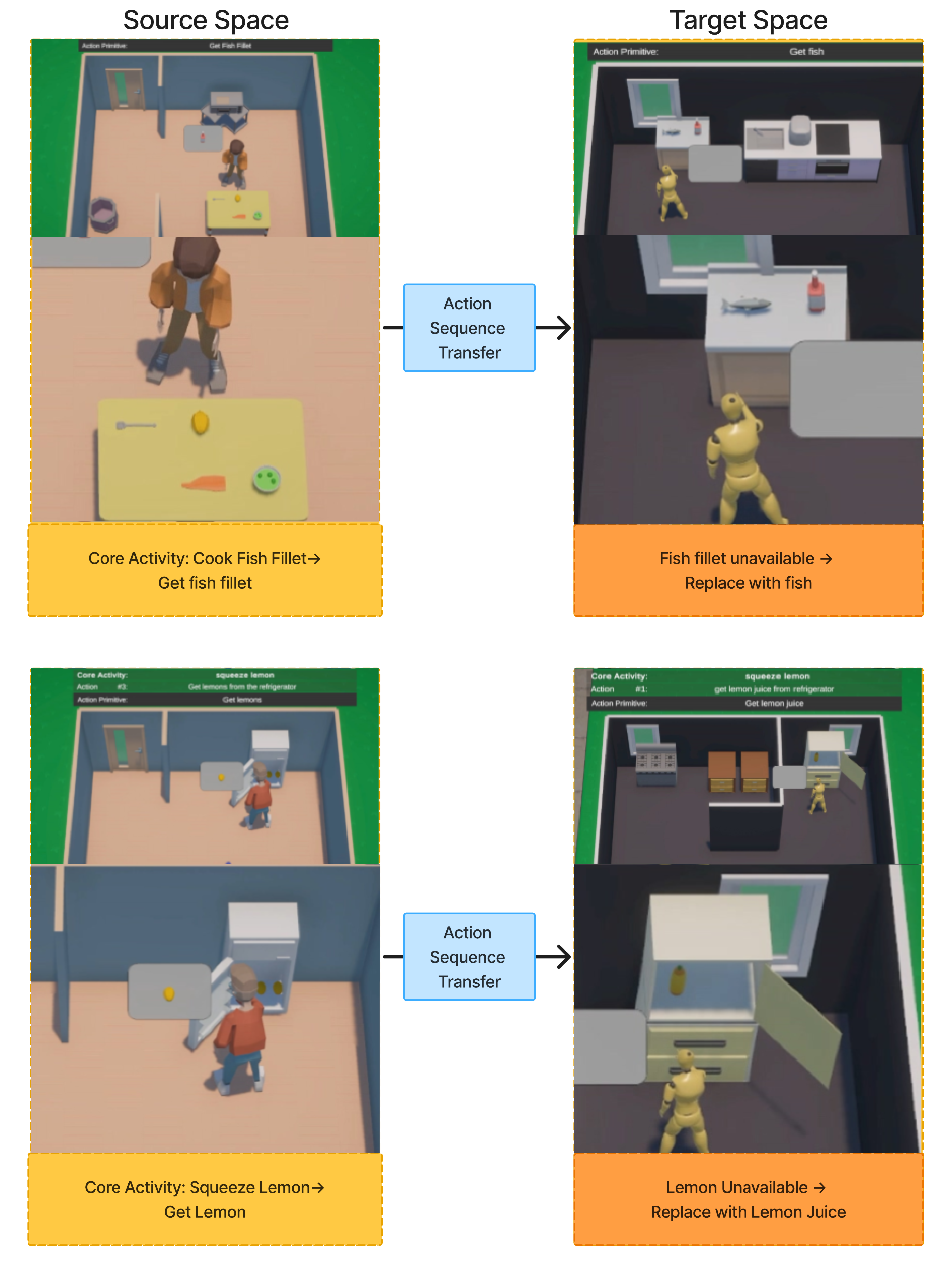}
\caption{Our framework enables robot in target space to actively use alternative objects to achieve activity goal, when corresponding objects are not available in the target environment.}
\label{fig:visualization_example}
\end{figure}

To illustrate the effectiveness of our framework, we present qualitative examples of action sequence transfer across heterogeneous environments as shown in Fig. \ref{fig:visualization_example}. In the source space, a human demonstrator cooks a fish fillet or squeezes a lemon. However, when these exact objects are unavailable in the target space, the framework enables the robot to adaptively substitute them with functionally similar alternatives while still achieving the intended activity goal. For instance, the absence of a fish fillet leads to a substitution with a whole fish, and the absence of a lemon results in the use of lemon juice instead.

These examples highlight the flexibility of our system in handling discrepancies between source and target environments. By relying on core activity and property representations, the framework can preserve the semantic intent of the original sequence while allowing substitutions that maintain task feasibility. Such adaptability demonstrates the practical potential of our approach in real-world robotic applications, where object inventories and spatial configurations often vary significantly across environments.

\subsection{Ablation Experiment}

\begin{figure}[t]
\centering
\includegraphics[width=\linewidth]{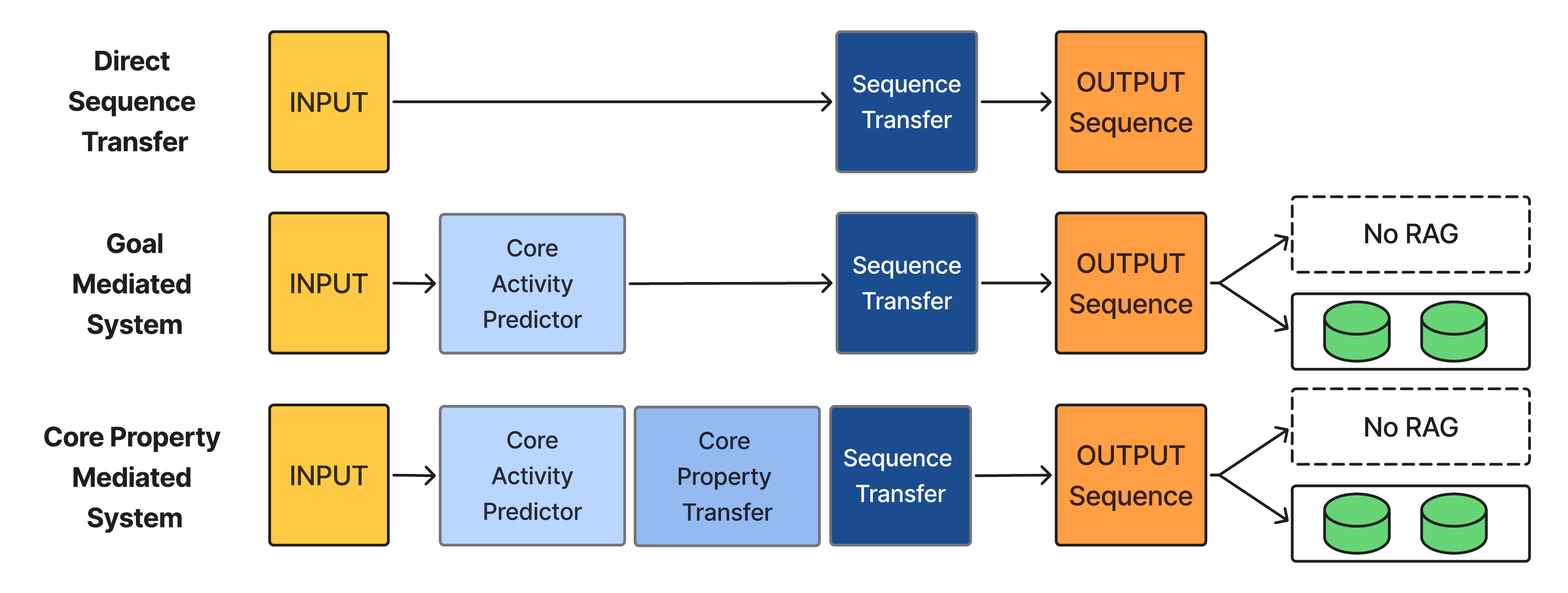}
\caption{We compare five architectures in our analysis: 1) direct sequence transfer architecture, 2) core activity architecture with and w/o RAG, and 3) core property architecture with and w/o RAG.}
\label{fig:eval_ablation}
\end{figure}

To evaluate the performance of our action sequence transfer system between two heterogeneous spaces, we examine two hypotheses concerning success rate 
and the effectiveness of external data retrieval for prompt augmentation.

\textbf{H1}: Leveraging fine-grained activity representation levels leads to a higher success rate of output action sequences for a given source-target space pair.


\textbf{H2}: Incorporating external data from similar activities for prompt augmentation improves the framework’s performance.

These hypotheses are evaluated using five framework variants (Fig. \ref{fig:eval_ablation}), grouped based on their use of multi-level user activity representations. The first group directly predicts the final action sequence without any abstraction or generalization. The second group uses only the predicted core activity to guide action sequence transfer. The third group incorporates both core activity and core property to provide fine-grained guidance.

\vspace{0.5em}
\textbf{Inference of target action sequence with evaluation dataset.}
Due to the ambiguity of goal description and goal category labels in the original Ego4d Goalstep dataset, we employ the core activity prediction model $\Phi_c$ to predict 600 core activities for each architecture, resulting in a total of 3,000 potential samples for acquiring ground truth core activity. The separate prediction of the core activity for different architectures has a dual purpose of examining the effect of RAG on valid goal prediction.

For each space pair sample, we manually examine both the initial scene graphs and the predicted core activity to exclude samples where execution of the predicted core activity is impossible (Tab. \ref{tab:success_rates}). This is due to the complete absence of key objects and viable substitutes in the target space. For direct transfer system that do not predict higher-level goals, we separately infer the core activity to apply the same feasibility filtering. This process yields 1,626 filtered cases where action sequence transfer to given target space is deemed possible.

The filtered cases are then used to generate final target action sequences for each architecture. The resulting sequences are evaluated by human to identify samples that both (1) make use of available objects and (2) follow appropriate procedures to accomplish the core activity. The success ratio of action transfer is defined as the proportion of successful samples among the filtered (feasible) input pairs.


\begin{figure}[t]
\centering
\includegraphics[width=1.0\linewidth]{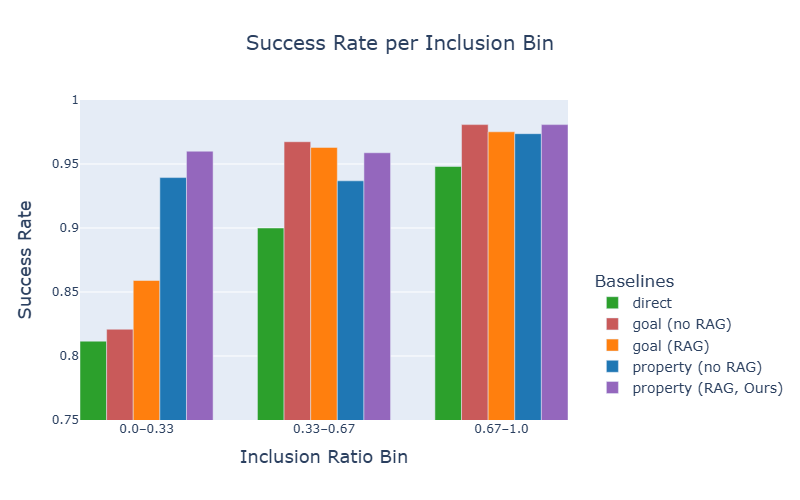}
\caption{For systems without fine-grained activity representations, success rates rapidly decrease for smaller scene inclusion ratios.}
\label{fig:eval_passratio}
\end{figure}

\vspace{0.5em}
\textbf{Success Ratio and Effect of RAG.} 
We evaluate the robustness of the action transfer system by plotting success rates across varying target scene inclusion ratios. The results reveal that the framework utilizing both core activity and core property representations achieves the highest success rate (Tab. \ref{tab:success_rates}). Additionally, incorporating RAG into core activity prediction significantly increases the number of validly filtered cases suitable for action sequence transfer, thereby improving the proportion of the test set that is deemed feasible.

\begin{table}[t]
\scriptsize
\centering
\renewcommand{\arraystretch}{1.2}
\setlength{\tabcolsep}{4pt}
\caption{Percentage of successful transfer across inclusion ratio ranges}
\label{tab:success_rates}
\begin{tabular}{|l|c|c|c|c|}
\hline
\multicolumn{5}{|c|}{\textit{Success Percentage for Filtered Samples}} \\
\hline
\textbf{Inclusion Ratio Range} & \textbf{0–0.33} & \textbf{0.33–0.67} & \textbf{0.67–1.0} & \textbf{Total (success / filtered)} \\
\hline
Direct & 81.2 & 90.0 & 94.8 & 90.4 (292 / 323) \\

Goal (no RAG) & 82.1 & \textbf{96.7} & \textbf{98.1} & 94.3 (297 / 315) \\

Goal (RAG) & 85.9 & 96.3 & 97.5 & 94.5 (328 / 347) \\

Property (no RAG) & 94.0 & 93.7 & 97.3 & 95.5 (299 / 313) \\

Property (RAG, \textbf{Ours}) & \textbf{96.0} & 95.9 & \textbf{98.1} & \textbf{97.0 (318 / 328)} \\
\hline
\end{tabular}
\end{table}

\begin{figure}[t]
\includegraphics[width=1.0\linewidth]{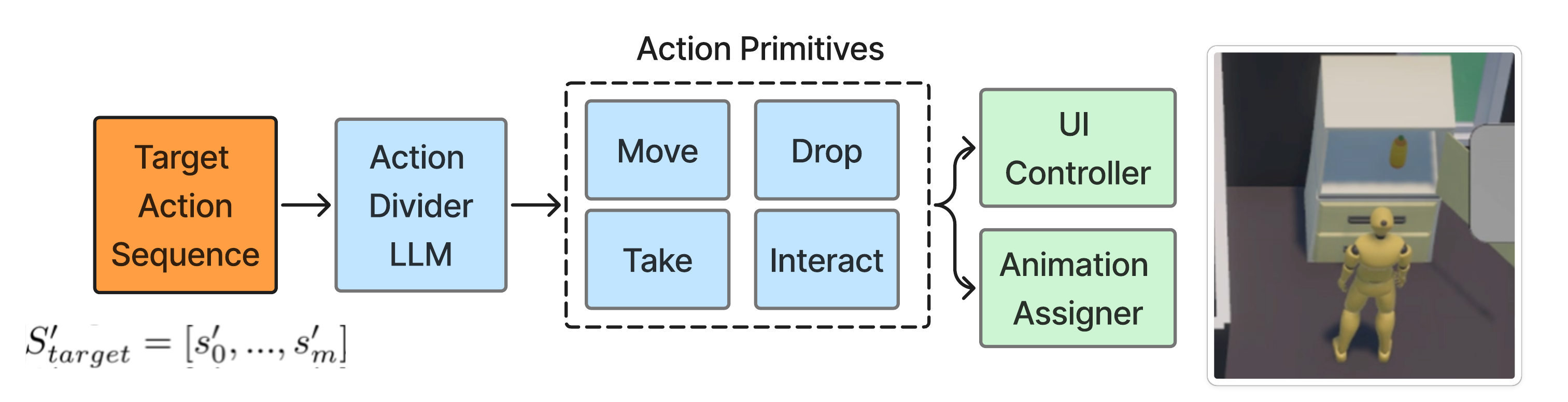}
\caption{In visualization, target action sequence is converted to action primitives for animating robot agent in target space.}
\label{fig:visualization_overview}
\end{figure}

\vspace{0.5em}
\textbf{Visualizing Execution of Action Sequences.}
We visualize the action sequences for both the source and target spaces within the Unity3D engine to demonstrate the robustness of our system in a virtual environment (Fig. \ref{fig:visualization_overview}). To initiate the visualization, we iterate through each action step in both sequences, further decomposing each step into four atomic action primitives—move, drop, take, and interact—using the Mistral 7B LLM \cite{jiang2024mixtral}. A UI controller displays the current action step, while an animation assigner triggers the corresponding recorded animation \cite{mixamo} for each action primitive.

Thanks to our exclusive focus on objects and action steps in spatial dataset annotation steps, our visualization system can address differences in geometric arrangements between the source and target space (Fig. \ref{fig:visualization_example}). Our action sequence transfer system allows for flexible alteration of key items, when identical objects are not available, enabling the robot in the target scene to execute slightly different tasks with different objects while still preserving the context of the core activity.

\section{DISCUSSION}

\subsection{Success Ratio Analysis}
The success rates across all systems suggest that incorporating granular descriptions to represent user action sequences significantly enhances the success rate of action sequence transfer, supporting H1. In simpler architectures—such as direct sequence transfer and core activity transfer—we observe a sharp decline in success rates, especially across spaces with low inclusion ratios. This indicates that simpler frameworks struggle to leverage alternative objects to reconstruct action sequences that bridge the differences between source and target spaces. This limitation underscores the value of incorporating more generalized, structured representations—such as core activities and core properties—into systems that employ multiple LLMs to procedurally identify functionally similar concepts or objects in the target space, even when exact matches from the source are unavailable.

\subsection{Retrieval Augmentation for Core Activity Inference}
Incorporating RAG into the core activity inference chain improves goal prediction accuracy and overall success rates, which supports H2.

The original Ego4D Goal-Step dataset provides both detailed multi-level action steps and their corresponding summarized goals, which supports more effective few-shot inference by the predictor LLM. Furthermore, augmenting the input with relevant spatial scene information—represented as partial scene graphs aligned with each action sub-step—adds essential temporal context about the types of actions and objects that are feasible within a given target space. 

In summary, the effectiveness of RAG in this setting stems from the close alignment between the structure of the external dataset and the formats of both the action sequences and the predicted core activities.


\subsection{Sequence Length Analysis}

\begin{table}[t]
\scriptsize
\centering
\resizebox{\columnwidth}{!}{%
\begin{tabular}{|l|c|c|c|c|c|}
\hline
\multicolumn{6}{|c|}{\textit{Action Count for a Successful Pair of Sequences}} \\
\hline
\textbf{}  & \textbf{Direct} & \textbf{Goal} & \textbf{Goal-RAG} & \textbf{Prop.} & \textbf{Prop.-RAG} \\
\hline
Source & 10.7 (7.6)  & 10.4 (7.4)  & 10.4 (7.1)  & 10.3 (7.2) & 10.3 (7.2)  \\
Target & 8.0 (6.2)  & 5.6 (5.0)  & 5.7 (4.8)  & 4.0 (3.2) & 4.1 (2.8)  \\
\hline
\end{tabular}
}
\caption{Mean and standard deviation of action counts in source action sequence and target action sequence for samples with successful action sequence transfer.}
\label{table:dataset_stats}
\end{table}

Table \ref{table:dataset_stats} shows the average length of source and target action sequences for success cases. We can see that conditioning LLMs on higher-level representations results in shorter target sequences. In contrast, simpler systems—such as direct sequence-to-sequence transfer or core-property-conditioned models with limited generative mechanisms—tend to preserve the structure of the source sequence by minimizing the removal of the original actions.

In the direct sequence transfer system, no constraints are imposed on preserving the core activity or property. As a result, when alternative objects are inferred during the action sequence transfer step, no comprehensive examination of the sequence occurs, leading to minimal refinement. This yields both lower success rates and higher preservation of the original sequence for successful cases.

Conversely, when preservation of core activity or core property is enforced during transfer, actions misaligned with higher-level intentions are more likely to be modified or removed. This is due to the presence of various sub-activities in the original dataset labeling and videos. Such minor actions, while not compromising the user's overall intent, can be excluded if they do not comply with the intended goals. This mechanism proves especially beneficial in low inclusion ratio scenarios, where a holistic understanding of user intent helps identify alternative strategies to achieve the goal. In such cases, complete restructuring of the action sequence based on higher-level directives can lead to successful results.

This analysis suggests a trade-off between structural preservation and semantic fidelity: systems without higher-level conditioning tend to retain more of the source sequence but often fail in challenging settings, whereas systems guided by abstract representations achieve higher success by producing more concise, goal-aligned sequences. These findings imply that effective sequence transfer may require balancing detail preservation with intentional simplification, depending on the application domain and the degree of heterogeneity between environments.

\section{CONCLUSION AND FUTURE WORKS}

In this paper, we present an action sequence transfer framework based on LLMs, designed to adapt user activities from a source space to fit the constraints of a different target space. Our results demonstrate that using structured representations of user activity enhances the system’s ability to adaptively modify and generate actions aligned with users’ higher-level goals.

While our framework ensures that the core activity and its key properties are preserved in the transferred sequences, we have not explicitly considered the preservation of the finer details of the source sequence in the target domain. Future work could focus on developing methods that strike a balance between semantic fidelity and detail preservation, so that the resulting target sequences not only achieve the overarching activity goal but also retain as much of the temporal and structural nuances of the original sequence as possible. We regard this as an important and challenging direction for advancing the robustness and expressiveness of sequence transfer systems.

Despite exploring various architectures, we still need to evaluate our framework across a broad range of open-source LLMs for better generalization. A follow-up comparative analysis of full fine-tuning versus few-shot inference could enable the development of lightweight, and efficient action transfer systems.

Finally, improvements to both the scale and quality of the dataset are necessary. While the current dataset supports the design of activity representations and external retrieval, it lacks direct visual grounding from the original Ego4D Goal-Step video data. Integrating a visual encoder or a vision-language model could significantly enhance the framework’s capabilities and broaden its applicability to visually rich, real-world environments.


\section*{ACKNOWLEDGMENT}
This work was supported by KOCCA, MCST (RS-2025-02307327) and IITP, MSIT (RS-2025-25441313).

                                  


\end{document}